\documentclass[11pt]{article}

\usepackage[final,nohyperref]{acl}

\usepackage{latexsym}
\usepackage[T2A,T1]{fontenc}
\usepackage[utf8]{inputenc}
\usepackage{times}
\usepackage{inconsolata}
\newcommand{\tj}[1]{{\fontencoding{T2A}\fontfamily{cmr}\selectfont #1}}
\DeclareUnicodeCharacter{04E2}{\={\CYRI}}
\DeclareUnicodeCharacter{04E3}{\={\cyri}}
\DeclareUnicodeCharacter{04EE}{\={\CYRU}}
\DeclareUnicodeCharacter{04EF}{\={\cyru}}
\usepackage{microtype}
\usepackage{graphicx}
\usepackage{booktabs}
\usepackage{enumitem}
\setlist{nosep,leftmargin=*}

\newcommand{\LfmBaseEnExact}{76.55}
\newcommand{\LfmBaseEnFn}{85.58}
\newcommand{\LfmBaseEnArgKey}{81.03}
\newcommand{\LfmBaseEnArgVal}{82.50}
\newcommand{\LfmBaseEnIrrel}{62.1}
\newcommand{\LfmBaseEnParse}{100.00}
\newcommand{\LfmBaseEnExtra}{4.14}
\newcommand{\LfmBaseEnSingleExact}{95.51}
\newcommand{\LfmBaseEnParallelExact}{70.37}
\newcommand{\LfmBaseEnMultiExact}{24.14}
\newcommand{\LfmBaseTgExact}{20.69}
\newcommand{\LfmBaseTgFn}{40.38}
\newcommand{\LfmBaseTgArgKey}{36.10}
\newcommand{\LfmBaseTgArgVal}{30.57}
\newcommand{\LfmBaseTgIrrel}{89.7}
\newcommand{\LfmBaseTgParse}{97.59}
\newcommand{\LfmBaseTgExtra}{7.59}
\newcommand{\LfmBaseTgSingleExact}{30.34}
\newcommand{\LfmBaseTgParallelExact}{7.41}
\newcommand{\LfmBaseTgMultiExact}{3.45}
\newcommand{\LfmCptOneEnExact}{0.00}
\newcommand{\LfmCptOneEnFn}{0.48}

\newcommand{\LfmCptOneTgExact}{0.00}
\newcommand{\LfmCptOneTgFn}{0.00}

\newcommand{\LfmCptTwoEnExact}{0.00}
\newcommand{\LfmCptTwoEnFn}{0.00}

\newcommand{\LfmCptTwoTgExact}{0.00}
\newcommand{\LfmCptTwoTgFn}{0.00}

\newcommand{\LfmSftOneEnExact}{77.24}
\newcommand{\LfmSftOneEnFn}{89.42}
\newcommand{\LfmSftOneEnArgKey}{83.61}
\newcommand{\LfmSftOneEnArgVal}{86.92}
\newcommand{\LfmSftOneEnIrrel}{31.0}
\newcommand{\LfmSftOneEnParse}{97.62}
\newcommand{\LfmSftOneEnExtra}{2.76}
\newcommand{\LfmSftOneEnSingleExact}{94.38}
\newcommand{\LfmSftOneEnParallelExact}{74.07}
\newcommand{\LfmSftOneEnMultiExact}{27.59}
\newcommand{\LfmSftOneTgExact}{64.14}
\newcommand{\LfmSftOneTgFn}{87.98}
\newcommand{\LfmSftOneTgArgKey}{82.87}
\newcommand{\LfmSftOneTgArgVal}{80.48}
\newcommand{\LfmSftOneTgIrrel}{6.9}
\newcommand{\LfmSftOneTgParse}{100.00}
\newcommand{\LfmSftOneTgExtra}{2.76}
\newcommand{\LfmSftOneTgSingleExact}{77.53}
\newcommand{\LfmSftOneTgParallelExact}{70.37}
\newcommand{\LfmSftOneTgMultiExact}{17.24}
\newcommand{\LfmSftTwoEnExact}{76.55}
\newcommand{\LfmSftTwoEnFn}{89.42}
\newcommand{\LfmSftTwoEnArgKey}{n/a}
\newcommand{\LfmSftTwoEnArgVal}{86.74}
\newcommand{\LfmSftTwoEnIrrel}{31.0}
\newcommand{\LfmSftTwoEnParse}{98.20}
\newcommand{\LfmSftTwoEnExtra}{2.76}

\newcommand{\LfmSftTwoTgExact}{63.45}
\newcommand{\LfmSftTwoTgFn}{88.46}
\newcommand{\LfmSftTwoTgArgKey}{n/a}
\newcommand{\LfmSftTwoTgArgVal}{80.29}
\newcommand{\LfmSftTwoTgIrrel}{10.3}
\newcommand{\LfmSftTwoTgParse}{100.00}
\newcommand{\LfmSftTwoTgExtra}{2.76}

\newcommand{\LfmNEnSingle}{89}
\newcommand{\LfmNEnParallel}{27}
\newcommand{\LfmNEnMulti}{29}
\newcommand{\LfmNEnIrrel}{29}

\newcommand{\LfmNAll}{174}
\newcommand{\LfmNCall}{145}
\newcommand{\LfmNIrrel}{29}
\newcommand{\LfmBaseEnExactLo}{69.7}
\newcommand{\LfmBaseEnExactHi}{83.4}
\newcommand{\LfmBaseEnSingleExactLo}{91.0}
\newcommand{\LfmBaseEnSingleExactHi}{98.9}
\newcommand{\LfmBaseEnParallelExactLo}{51.9}
\newcommand{\LfmBaseEnParallelExactHi}{85.2}
\newcommand{\LfmBaseEnMultiExactLo}{10.3}
\newcommand{\LfmBaseEnMultiExactHi}{41.4}
\newcommand{\LfmBaseTgExactLo}{14.5}
\newcommand{\LfmBaseTgExactHi}{26.9}
\newcommand{\LfmBaseTgSingleExactLo}{21.3}
\newcommand{\LfmBaseTgSingleExactHi}{40.4}
\newcommand{\LfmBaseTgParallelExactLo}{0.0}
\newcommand{\LfmBaseTgParallelExactHi}{18.5}
\newcommand{\LfmBaseTgMultiExactLo}{0.0}
\newcommand{\LfmBaseTgMultiExactHi}{10.3}
\newcommand{\LfmCptOneEnExactLo}{0.0}
\newcommand{\LfmCptOneEnExactHi}{0.0}
\newcommand{\LfmCptOneTgExactLo}{0.0}
\newcommand{\LfmCptOneTgExactHi}{0.0}
\newcommand{\LfmCptTwoEnExactLo}{0.0}
\newcommand{\LfmCptTwoEnExactHi}{0.0}
\newcommand{\LfmCptTwoTgExactLo}{0.0}
\newcommand{\LfmCptTwoTgExactHi}{0.0}
\newcommand{\LfmSftOneEnExactLo}{70.3}
\newcommand{\LfmSftOneEnExactHi}{84.1}
\newcommand{\LfmSftOneEnSingleExactLo}{88.8}
\newcommand{\LfmSftOneEnSingleExactHi}{98.9}
\newcommand{\LfmSftOneEnParallelExactLo}{55.6}
\newcommand{\LfmSftOneEnParallelExactHi}{88.9}
\newcommand{\LfmSftOneEnMultiExactLo}{10.3}
\newcommand{\LfmSftOneEnMultiExactHi}{44.8}
\newcommand{\LfmSftOneTgExactLo}{55.9}
\newcommand{\LfmSftOneTgExactHi}{71.7}
\newcommand{\LfmSftOneTgSingleExactLo}{68.5}
\newcommand{\LfmSftOneTgSingleExactHi}{85.4}
\newcommand{\LfmSftOneTgParallelExactLo}{51.9}
\newcommand{\LfmSftOneTgParallelExactHi}{85.2}
\newcommand{\LfmSftOneTgMultiExactLo}{3.4}
\newcommand{\LfmSftOneTgMultiExactHi}{31.0}
\newcommand{\LfmSftTwoEnExactLo}{69.7}
\newcommand{\LfmSftTwoEnExactHi}{83.4}
\newcommand{\LfmSftTwoTgExactLo}{55.2}
\newcommand{\LfmSftTwoTgExactHi}{71.0}
\newcommand{\LfmPairEnBaseSftDelta}{\ensuremath{+0.69}}
\newcommand{\LfmPairEnBaseSftLo}{\ensuremath{-4.1}}
\newcommand{\LfmPairEnBaseSftHi}{\ensuremath{+4.8}}

\newcommand{\LfmPairTgBaseSftDelta}{\ensuremath{+43.45}}
\newcommand{\LfmPairTgBaseSftLo}{\ensuremath{+35.2}}
\newcommand{\LfmPairTgBaseSftHi}{\ensuremath{+51.7}}
\newcommand{\LfmPairTgBaseSftOnlyA}{2}
\newcommand{\LfmPairTgBaseSftOnlyB}{65}
\newcommand{\LfmPairTgBaseSftP}{\ensuremath{<10^{-4}}}
\newcommand{\LfmPairEnSftOneTwoDelta}{\ensuremath{-0.69}}

\newcommand{\LfmPairTgSftOneTwoDelta}{\ensuremath{-0.69}}

\newcommand{\LfmPairTgSftOneTwoOnlyA}{3}
\newcommand{\LfmPairTgSftOneTwoOnlyB}{2}
\newcommand{\LfmPairTgSftOneTwoP}{1.0000}

\newcommand{\LfmPairEnIrrelBaseSftP}{0.0352}

\newcommand{\LfmBaseEnTagCount}{154}
\newcommand{\LfmBaseEnJsonCount}{0}
\newcommand{\LfmBaseEnNeitherCount}{20}
\newcommand{\LfmBaseTgTagCount}{83}
\newcommand{\LfmBaseTgJsonCount}{12}
\newcommand{\LfmBaseTgNeitherCount}{79}
\newcommand{\LfmCptOneEnTagCount}{1}
\newcommand{\LfmCptOneEnJsonCount}{107}
\newcommand{\LfmCptOneEnNeitherCount}{66}
\newcommand{\LfmCptOneTgTagCount}{0}
\newcommand{\LfmCptOneTgJsonCount}{52}
\newcommand{\LfmCptOneTgNeitherCount}{122}
\newcommand{\LfmCptTwoEnTagCount}{0}
\newcommand{\LfmCptTwoEnJsonCount}{60}
\newcommand{\LfmCptTwoEnNeitherCount}{114}
\newcommand{\LfmCptTwoTgTagCount}{0}
\newcommand{\LfmCptTwoTgJsonCount}{37}
\newcommand{\LfmCptTwoTgNeitherCount}{137}
\newcommand{\LfmSftOneEnTagCount}{168}
\newcommand{\LfmSftOneEnJsonCount}{0}
\newcommand{\LfmSftOneEnNeitherCount}{6}
\newcommand{\LfmSftOneTgTagCount}{171}
\newcommand{\LfmSftOneTgJsonCount}{0}
\newcommand{\LfmSftOneTgNeitherCount}{3}
\newcommand{\LfmCptSteps}{17,574}
\newcommand{\LfmCptEpochs}{3}
\newcommand{\LfmCptGpus}{3}

\newcommand{\LfmCptRuntimeHours}{47.1}
\newcommand{\LfmCptTrainLoss}{1.888}
\newcommand{\LfmCptLossFirst}{6.615}
\newcommand{\LfmCptLossLast}{1.662}

\newcommand{\LfmCptLr}{2e-5}
\newcommand{\LfmCptSched}{cosine}
\newcommand{\LfmCptWarmup}{0.03}
\newcommand{\LfmCptClip}{1.0}
\newcommand{\LfmCptLoraR}{32}
\newcommand{\LfmCptLoraAlpha}{32}

\newcommand{\LfmCptSeqLen}{4,096}
\newcommand{\LfmCptMicroBatch}{8}
\newcommand{\LfmCptGradAccum}{4}
\newcommand{\LfmCptEffBatch}{96}

\newcommand{\LfmSftOneSteps}{3,678}
\newcommand{\LfmSftOneLossFirst}{0.879}
\newcommand{\LfmSftOneLossLast}{0.244}
\newcommand{\LfmSftOneTrainLoss}{0.309}

\newcommand{\LfmSftOneRuntimeHours}{2.11}
\newcommand{\LfmSftOneLr}{2e-5}
\newcommand{\LfmSftOneEpochs}{1}

\newcommand{\LfmSftTwoSteps}{3,678}
\newcommand{\LfmSftTwoLossFirst}{0.280}
\newcommand{\LfmSftTwoLossLast}{0.123}
\newcommand{\LfmSftTwoTrainLoss}{0.261}

\newcommand{\LfmSftTwoRuntimeHours}{2.13}
\newcommand{\LfmSftTwoLr}{1e-5}
\newcommand{\LfmSftTwoEpochs}{1}

\newcommand{\LfmSftMicroBatch}{2}
\newcommand{\LfmSftGradAccum}{8}
\newcommand{\LfmSftGpus}{3}
\newcommand{\LfmSftEffBatch}{48}
\newcommand{\LfmSftSeqLen}{4,096}
\newcommand{\LfmSftClip}{0.5}
\newcommand{\LfmSftLoraR}{32}
\newcommand{\LfmSftLoraAlpha}{32}
\newcommand{\LfmSftWarmup}{0.03}

\newcommand{\LfmSftTwoLrTex}{1e-5}

\newcommand{\LfmDocsInput}{1,615,358}

\newcommand{\LfmDocsPre}{1,615,356}
\newcommand{\LfmDocsKept}{1,485,319}
\newcommand{\LfmDroppedResidual}{129,990}
\newcommand{\LfmDroppedDup}{49}
\newcommand{\LfmDocsKeptPct}{91.9}

\newcommand{\LfmTokensPre}{2,433,790,558}

\newcommand{\LfmTokensPreB}{2.43}
\newcommand{\LfmSrcDocsHplt}{441,838}
\newcommand{\LfmSrcDocsFineweb}{447,146}
\newcommand{\LfmSrcDocsMadlad}{276,211}
\newcommand{\LfmSrcDocsCfour}{217,081}
\newcommand{\LfmSrcDocsWiki}{50,024}
\newcommand{\LfmSrcDocsShah}{8}
\newcommand{\LfmSrcDocsPretrain}{183,048}
\newcommand{\LfmSrcDocsSourceOne}{1,432,308}

\newcommand{\LfmSrcPctHplt}{27.4}
\newcommand{\LfmSrcPctFineweb}{27.7}
\newcommand{\LfmSrcPctMadlad}{17.1}
\newcommand{\LfmSrcPctCfour}{13.4}
\newcommand{\LfmSrcPctWiki}{3.1}
\newcommand{\LfmSrcPctShah}{0.0}
\newcommand{\LfmSrcPctPretrain}{11.3}
\newcommand{\LfmToolRowsGlaive}{54,859}

\newcommand{\LfmToolRowsHermesFc}{1,090}

\newcommand{\LfmToolRowsHermesSingle}{1,090}

\newcommand{\LfmToolRowsHermesGlaive}{4,342}

\newcommand{\LfmToolRowsToolAce}{9,245}

\newcommand{\LfmToolRowsTotal}{70,626}
\newcommand{\LfmToolDupDropped}{8,334}

\newcommand{\LfmSftMixRows}{176,565}
\newcommand{\LfmSftPretokRows}{176,532}
\newcommand{\LfmSftPretokDropped}{33}
\newcommand{\LfmMixTgTool}{70,626}
\newcommand{\LfmMixEnTool}{61,798}
\newcommand{\LfmMixInstruct}{44,141}
\newcommand{\LfmMixTgPct}{40}
\newcommand{\LfmMixEnPct}{35}
\newcommand{\LfmMixInsPct}{25}

\newcommand{\LfmSurfaceStrings}{77,453}
\newcommand{\LfmTranslateModel}{NLLB-200 distilled 1.3B}

\newcommand{\LfmCptPacking}{true}
\newcommand{\LfmSftPacking}{false}

\newcommand{\LfmSftSched}{cosine}
\newcommand{\LfmLenBaseEnExact}{76.55}
\newcommand{\LfmLenBaseEnFn}{85.58}

\newcommand{\LfmInfBaseEnExact}{76.55}

\newcommand{\LfmLenBaseTgExact}{31.72}
\newcommand{\LfmLenBaseTgFn}{61.54}

\newcommand{\LfmLenBaseTgIrrel}{31.0}

\newcommand{\LfmInfBaseTgExact}{32.41}

\newcommand{\LfmLenCptOneEnExact}{35.17}
\newcommand{\LfmLenCptOneEnFn}{34.62}

\newcommand{\LfmLenCptOneEnMultiExact}{6.90}
\newcommand{\LfmInfCptOneEnExact}{48.28}

\newcommand{\LfmLenCptOneTgExact}{6.90}
\newcommand{\LfmLenCptOneTgFn}{10.58}

\newcommand{\LfmLenCptOneTgMultiExact}{0.00}
\newcommand{\LfmInfCptOneTgExact}{15.17}

\newcommand{\LfmLenCptTwoEnExact}{10.34}
\newcommand{\LfmLenCptTwoEnFn}{9.13}

\newcommand{\LfmInfCptTwoEnExact}{20.00}

\newcommand{\LfmLenCptTwoTgExact}{6.90}
\newcommand{\LfmLenCptTwoTgFn}{9.13}

\newcommand{\LfmInfCptTwoTgExact}{9.66}

\newcommand{\LfmLenSftOneEnExact}{77.24}
\newcommand{\LfmLenSftOneEnFn}{89.42}

\newcommand{\LfmInfSftOneEnExact}{77.24}

\newcommand{\LfmLenSftOneTgExact}{64.14}
\newcommand{\LfmLenSftOneTgFn}{87.98}

\newcommand{\LfmInfSftOneTgExact}{64.14}

\newcommand{\LfmMcqBaseBel}{37.6}
\newcommand{\LfmMcqBaseBelLo}{34.4}
\newcommand{\LfmMcqBaseBelHi}{40.8}
\newcommand{\LfmMcqCptFinBel}{36.6}
\newcommand{\LfmMcqCptFinBelLo}{33.4}
\newcommand{\LfmMcqCptFinBelHi}{39.7}
\newcommand{\LfmMcqSftOneBel}{40.0}
\newcommand{\LfmMcqSftOneBelLo}{36.8}
\newcommand{\LfmMcqSftOneBelHi}{43.2}
\newcommand{\LfmMcqSftTwoBel}{40.9}
\newcommand{\LfmMcqSftTwoBelLo}{37.8}
\newcommand{\LfmMcqSftTwoBelHi}{44.0}
\newcommand{\LfmNBel}{900}
\newcommand{\LfmMcqBaseCur}{28.0}
\newcommand{\LfmMcqBaseCurLo}{24.7}
\newcommand{\LfmMcqBaseCurHi}{31.5}
\newcommand{\LfmMcqCptFinCur}{33.0}
\newcommand{\LfmMcqCptFinCurLo}{29.5}
\newcommand{\LfmMcqCptFinCurHi}{36.7}
\newcommand{\LfmMcqSftOneCur}{35.0}
\newcommand{\LfmMcqSftOneCurLo}{31.5}
\newcommand{\LfmMcqSftOneCurHi}{38.7}
\newcommand{\LfmMcqSftTwoCur}{35.0}
\newcommand{\LfmMcqSftTwoCurLo}{31.4}
\newcommand{\LfmMcqSftTwoCurHi}{38.7}
\newcommand{\LfmNCur}{657}
\newcommand{\LfmMcqBaseMmlu}{60.8}
\newcommand{\LfmMcqBaseMmluLo}{60.0}
\newcommand{\LfmMcqBaseMmluHi}{61.6}
\newcommand{\LfmMcqCptFinMmlu}{49.5}
\newcommand{\LfmMcqCptFinMmluLo}{48.7}
\newcommand{\LfmMcqCptFinMmluHi}{50.3}
\newcommand{\LfmMcqSftOneMmlu}{52.5}
\newcommand{\LfmMcqSftOneMmluLo}{51.7}
\newcommand{\LfmMcqSftOneMmluHi}{53.3}
\newcommand{\LfmMcqSftTwoMmlu}{52.5}
\newcommand{\LfmMcqSftTwoMmluLo}{51.7}
\newcommand{\LfmMcqSftTwoMmluHi}{53.4}
\newcommand{\LfmNMmlu}{14,042}

\newcommand{\LfmMcqPairBelBaseSftDelta}{\ensuremath{+2.4}}
\newcommand{\LfmMcqPairBelBaseSftLo}{\ensuremath{-1.4}}
\newcommand{\LfmMcqPairBelBaseSftHi}{\ensuremath{+6.3}}

\newcommand{\LfmMcqPairCurBaseSftDelta}{\ensuremath{+7.0}}
\newcommand{\LfmMcqPairCurBaseSftLo}{\ensuremath{+2.4}}
\newcommand{\LfmMcqPairCurBaseSftHi}{\ensuremath{+11.7}}
\newcommand{\LfmMcqPairCurBaseSftP}{0.0039}

\newcommand{\LfmMcqPairMmluBaseCptDelta}{\ensuremath{-11.3}}

\newcommand{\LfmZeroUpper}{2.5}
\newcommand{\LfmMcqSftOneBelAboveChance}{15}
\newcommand{\LfmMcqSftOneCurAboveChance}{10}
\newcommand{\LfmTgUntranslated}{16}
\newcommand{\LfmTgUntranslatedParallel}{13}

\newcommand{\LfmGapEnTgSft}{13.10}

\newcommand{\LfmGlaiveToolsOffered}{78,362}
\newcommand{\LfmGlaiveNoCall}{15,144}
\newcommand{\LfmToolCptFinEnExact}{0.00}

\newcommand{\LfmToolCptFinEnTagCount}{0}

\newcommand{\LfmToolCptFinTgTagCount}{0}
\newcommand{\LfmToolRerunEnExact}{75.86}

\newcommand{\LfmRerunEnSame}{167}
\newcommand{\LfmRerunEnDiff}{7}
\newcommand{\LfmRerunEnShift}{1.38}
\newcommand{\LfmToolRerunTgExact}{64.14}

\newcommand{\LfmRerunTgSame}{168}

\newcommand{\LfmTokVocab}{125,017}
\newcommand{\LfmTokTgCpt}{2.36}
\newcommand{\LfmTokEnCpt}{4.32}
\newcommand{\LfmTokTgTpw}{2.61}
\newcommand{\LfmTokEnTpw}{1.30}
\newcommand{\LfmTokTurns}{220}
\newcommand{\ModelName}{Saidzoda-Tajik-LFM-2.6B}
\newcommand{\LfmGarbledLo}{53,662}
\newcommand{\LfmGarbledHi}{85,970}
\newcommand{\LfmGarbledLoPct}{3.6}
\newcommand{\LfmGarbledHiPct}{5.8}
\newcommand{\LfmQwenSameIds}{281}
\newcommand{\LfmQwenIdsChecked}{151,643}
\newcommand{\LfmLenHandN}{20}
\newcommand{\LfmLenHandTrue}{20}

\title{\ModelName: Adapting a 2.6B Hybrid Model\\to Tajik Tool Calling, and What Continued Pretraining Takes Away}

\author{Tohir Saidzoda \and Nekruz Maxudov \and Bedil Karimov \\
  \textbf{Abubakr Abdullozoda} \and \textbf{Amin Jalilov} \and \textbf{Jonibek Kishvarov} \\
  \textbf{Rahmonali Sharipov} \and \textbf{Mariyamkhon Shodyeva} \\
  Saidzoda AI Lab, Saidzoda Engineering Company, Dushanbe, Tajikistan \\
  \texttt{tohir@saidzoda.com}}

\begin{document}
\maketitle

\begin{abstract}
We release \ModelName, a Tajik adaptation of the 2.6B-parameter hybrid model LFM2.5-2.6B-Base, built for tool calling in Tajik on modest hardware.
The recipe has two stages: LoRA continued pretraining (CPT) on \LfmDocsKept{} cleaned Tajik web documents for \LfmCptEpochs{} epochs, then one epoch of supervised fine-tuning (SFT) on a mix of machine-translated Tajik tool calls, English tool calls kept as replay, and Tajik instructions.
On a \LfmNAll{}-item tool-calling test set built from BFCL v3 and machine-translated to Tajik, strict full-call accuracy on Tajik prompts rises from \LfmBaseTgExact{} for the base model to \LfmSftOneTgExact{} (paired difference \LfmPairTgBaseSftDelta{} points, 95\% interval \LfmPairTgBaseSftLo{} to \LfmPairTgBaseSftHi{}); a lenient parser that also accepts the base model's own call formats puts the base at \LfmLenBaseTgExact.
English shows no detectable change (\LfmBaseEnExact{} and \LfmSftOneEnExact).
Every CPT checkpoint scores zero: CPT removes the native tool-call format and, as the lenient re-scoring shows, part of the skill itself.
Outside tool calling, accuracy on a Tajik knowledge benchmark rises from \LfmMcqBaseCur{} to \LfmMcqSftOneCur, Belebele Tajik shows no significant change, and English MMLU falls from \LfmMcqBaseMmlu{} to \LfmMcqSftOneMmlu.
The model is not a general Tajik assistant.
It almost never declines when no tool fits (\LfmSftOneTgIrrel\% correct on Tajik, against \LfmLenBaseTgIrrel\% for the base under the lenient parser), it is weak at multi-turn calls, and its calls should be checked before they are executed.
We describe the data, the training settings and every evaluation in enough detail to repeat them.

\end{abstract}

\section{Introduction}

Tajik, the state language of Tajikistan, is written in a Cyrillic alphabet with six letters that Russian does not have (\tj{ғ, ӣ, қ, ӯ, ҳ, ҷ}).
Open language models rarely list it among their supported languages, and the few Tajik models that exist are large and aimed at chat and knowledge questions \citep{liashkov2026soro}.
Many practical uses in Tajikistan, such as call centres, government services and banking, need something different: a small model that can run on modest hardware and turn a Tajik request into a correct call to an existing system.

Small open models now do this well in English.
LFM2.5-2.6B, a hybrid of short convolutions and grouped-query attention from Liquid AI \citep{amini2025lfm2}, answers \LfmBaseEnExact{}\% of our English tool-calling items exactly, but only \LfmBaseTgExact{}\% when the same request is written in Tajik.
The obvious fix, continued pretraining on Tajik text followed by fine-tuning, carries a known risk: training on raw text can remove skills the model learned in post-training \citep{biderman2024lora}.

This report describes how we built \ModelName{} and what we measured along the way.
Our contributions are:
\begin{enumerate}
\item \textbf{A released model.} To our knowledge the first public Tajik tool-calling model released with a technical report, at 2.6B parameters, with the full recipe: data sources and cleaning, configurations, compute, and training logs (Section~\ref{sec:recipe}).
\item \textbf{A Tajik tool-calling test set and scorer.} \LfmNAll{} items in four categories from BFCL v3 \citep{patil2025bfcl}, with the user turns machine-translated to Tajik and the tools and gold calls unchanged, scored by an AST matcher (Section~\ref{sec:eval}).
\item \textbf{A stage-by-stage measurement.} Base, two CPT checkpoints and two SFT epochs, in both languages, with bootstrap intervals and paired tests (Section~\ref{sec:results}). Raw-text CPT takes the strict score to zero; SFT with English replay restores English and lifts Tajik well above the base.
\item \textbf{Format or skill.} A lenient re-scoring of the same outputs separates the loss of the output format from the loss of the skill itself (Section~\ref{sec:format}).
\item \textbf{Known weaknesses.} Abstention and multi-turn calls are weak after SFT, and a second epoch does not help. We report these as results (Section~\ref{sec:weak}).
\end{enumerate}

\section{Related Work}

\paragraph{Adapting models to new languages.}
Continued pretraining on target-language text, often with vocabulary extension, is the standard way to add a language to an existing model \citep{cui2023chinesellama}, and the multilingual pretraining literature shows that languages compete for capacity \citep{chang2023curse}.
Continued pretraining also causes forgetting \citep{kirkpatrick2017ewc,luo2023forgetting}; rewarming the learning rate and replaying earlier data reduce it \citep{gupta2023rewarm,ibrahim2024simple}.
LoRA \citep{hu2021lora} and its rank-stabilised form \citep{kalajdzievski2023rslora} make such training cheap and forget less than full fine-tuning, but also learn less \citep{biderman2024lora}.
Our report adds a measurement of what this costs for one specific skill, tool calling, on a small hybrid model.

\paragraph{Tool calling.}
Language models learn to call tools from self-supervised traces \citep{schick2023toolformer}, from API documentation \citep{patil2023gorilla}, and from large synthetic sets such as ToolACE, APIGen and xLAM \citep{liu2024toolace,liu2024apigen,zhang2024xlam}.
Small models can be competitive when trained for the task \citep{lin2024hammer,erdogan2024tinyagent}.
BFCL \citep{patil2025bfcl} is the most common benchmark; we build on its test files.
Fine-tuning a small model on function calls can also cost general ability, and mixing in general data reduces the loss \citep{ran2024alopex}; we see the mirror image, where general-text training costs tool calling.

\paragraph{Tool calling beyond English.}
Recent multilingual benchmarks and training sets cover several languages but not Tajik \citep{huang2025benchmax,zhang2026itc}.
Two findings matter directly for us.
First, when BFCL is translated, models still pick the right tool but often fill parameters with values in the user's language where the API expects language-independent ones \citep{luo2026lostinexecution,chauhan2026apispeaks}; the latter also find plain SFT a strong fix.
Second, TUCAN adapted a Bulgarian model to function calling with bilingual SFT, the same overall recipe as ours, on a base that already spoke the language well \citep{emanuilov2025tucan}.
Our setting differs in that the base speaks almost no Tajik and must first learn the language through CPT.

\paragraph{Tajik.}
Tajik has few open resources.
Recent work has released Tajik web text and dictionary resources \citep{arabov2026tajikweb,arabov2026dictionary,arabov2026tajiknlp}, and larger Tajik chat models have been released \citep{liashkov2026soro}.
Belebele covers Tajik reading comprehension \citep{bandarkar2024belebele}.
Our web data comes from multilingual crawls with Tajik subsets \citep{burchell2025hplt2,penedo2025fineweb2,kudugunta2023madlad400,raffel2020t5}.
To our knowledge \ModelName{} is the first public Tajik tool-calling model released with a technical report, and our test set is the first Tajik tool-calling test set; we searched arXiv and the Hugging Face Hub for Tajik tool-calling and function-calling models and datasets in October 2026 and found none.

\section{Model and Training Recipe}
\label{sec:recipe}

\subsection{Base model and tokenizer}
We start from LFM2.5-2.6B-Base \citep{amini2025lfm2}, the pretrained (not instruction-tuned) release.
We confirmed the identity of the weights we trained on by comparing the SHA-256 hash of our copy with the file on the Hugging Face Hub.
Before training we checked that the tokenizer can represent Tajik at all, since a missing letter cannot be learned from more data.
The vocabulary has \LfmTokVocab{} entries, and each of the six Tajik-specific lowercase letters is a single token.
On the user turns of our test set (\LfmTokTurns{} turns per language), Tajik text averages \LfmTokTgCpt{} characters per token against \LfmTokEnCpt{} for the English originals, or \LfmTokTgTpw{} against \LfmTokEnTpw{} tokens per word.
Tajik is therefore about twice as expensive per word, but no change to the vocabulary or embeddings was needed.

\subsection{Continued pretraining data}
The CPT corpus combines public Tajik web text with a Tajik pretraining set of our own (Table~\ref{tab:lfm-cptdata} in the appendix).
The public part takes the Tajik subsets of HPLT 2.0, FineWeb-2 \citep{penedo2025fineweb2}, MADLAD-400, C4 and a Tajik Wikipedia dump, \LfmSrcDocsSourceOne{} documents in total; our own set adds \LfmSrcDocsPretrain{} sequences.
After exact deduplication the corpus held \LfmDocsPre{} documents and \LfmTokensPreB{}B tokens under the LFM2.5 tokenizer.

A share of Tajik text on the web was typed with legacy fonts that map Tajik letters onto other Cyrillic code points (for example \tj{ќ} for \tj{қ} and \tj{љ} for \tj{ҷ}), so the same word appears in several spellings.
We mapped the known substitutions to the correct letters and then dropped every document with fewer than 20 Cyrillic characters or with more than 1\% of its Cyrillic characters outside the Tajik and Russian alphabets.
After a second, normalised deduplication this removed \LfmDroppedResidual{} documents and \LfmDroppedDup{} further duplicates, keeping \LfmDocsKept{} documents (\LfmDocsKeptPct\%).

\paragraph{A defect we found while writing this report.}
Our own pretraining set is stored as token ids from a Qwen2.5 tokenizer, but the cleaning step decoded it with the Qwen3.5 tokenizer.
The two vocabularies agree on only \LfmQwenSameIds{} of the first \LfmQwenIdsChecked{} ids, so the decoded text is mostly a meaningless mix of tokens from other scripts.
The filter above counts only Cyrillic characters and removed most of these rows, but not all of them: from the counters in the cleaning log, between \LfmGarbledLo{} and \LfmGarbledHi{} of the kept documents (\LfmGarbledLoPct{} to \LfmGarbledHiPct\%) come from this set.
The model was trained on the corpus as it was, and every result below is measured on that model; we report the defect so that the recipe can be repeated correctly.

\subsection{Continued pretraining}
\begin{figure}[t]
\centering
\includegraphics[width=\columnwidth]{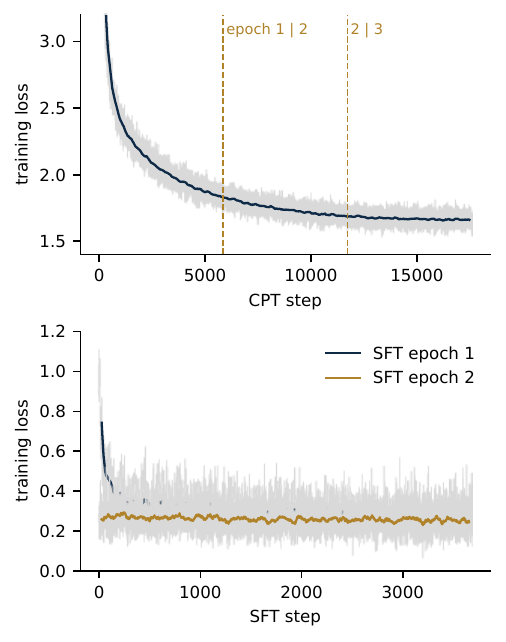}
\caption{Training loss for CPT (top, epoch boundaries marked) and for the two SFT epochs (bottom). Light lines are per-step values, dark lines a moving average.}
\label{fig:loss}
\end{figure}

CPT trains a LoRA adapter \citep{hu2021lora} with rank-stabilised scaling, which divides the LoRA scale by the square root of the rank \citep{kalajdzievski2023rslora}, on raw text as plain language modelling.
The adapter covers the attention projections, the convolution input and output projections and the feed-forward layers (Table~\ref{tab:lfm-recipe}).
We trained for \LfmCptEpochs{} epochs, \LfmCptSteps{} optimiser steps with \LfmCptEffBatch{} packed sequences of \LfmCptSeqLen{} tokens each, on \LfmCptGpus{} H100 GPUs in \LfmCptRuntimeHours{} hours.
The training loss fell from \LfmCptLossFirst{} at the first step to \LfmCptLossLast{} at the last, with no spikes (Figure~\ref{fig:loss}).
The adapter was then merged into the base weights.

\subsection{Supervised fine-tuning data}
The SFT mix (Table~\ref{tab:lfm-sftdata} in the appendix) has three parts, chosen before training at \LfmMixTgPct{}/\LfmMixEnPct{}/\LfmMixInsPct{} percent:
\begin{itemize}
\item \textbf{Tajik tool calls (\LfmMixTgTool{} rows).} English tool-calling conversations from Glaive function-calling v2, Hermes function-calling v1 and ToolACE \citep{liu2024toolace}, deduplicated (\LfmToolDupDropped{} exact duplicates removed). The natural-language text of each conversation, \LfmSurfaceStrings{} strings in all, was translated to Tajik with \LfmTranslateModel{} \citep{nllb2022}. Quoted strings, URLs and e-mail addresses were masked before translation and restored after it. Tool names, schemas and argument keys stay in English, as they would in a real API.
\item \textbf{English replay (\LfmMixEnTool{} rows).} The same English tool-calling data, untranslated, sampled at random. Its purpose is to rebuild the native call format and keep the English skill.
\item \textbf{Tajik instructions (\LfmMixInstruct{} rows).} General Tajik instruction data from our own SFT corpus, so that the model keeps answering in Tajik when no tool is involved.
\end{itemize}
Every conversation was rendered with the model's own chat template, including tool definitions, and tokenised offline with loss on assistant tokens only.
\LfmSftPretokDropped{} rows longer than \LfmSftSeqLen{} tokens were dropped, leaving \LfmSftPretokRows{}.

\subsection{Supervised fine-tuning}
SFT starts from the merged CPT model and trains a new LoRA adapter with the same targets for one epoch, \LfmSftOneSteps{} steps at an effective batch of \LfmSftEffBatch{} conversations, in \LfmSftOneRuntimeHours{} hours on \LfmSftGpus{} GPUs.
The loss started at \LfmSftOneLossFirst{} and settled near \LfmSftOneLossLast{}.
The merged result is the released model.
To test whether more training would help, we trained a second epoch on the same data at half the learning rate (\LfmSftTwoLrTex{}) starting from the merged epoch-1 model.

\section{Evaluation}
\label{sec:eval}

\paragraph{Test set.}
We built a fixed test set of \LfmNAll{} items from the BFCL v3 test files \citep{patil2025bfcl} with a seeded script: \LfmNEnSingle{} \emph{single} items (one tool, one call), \LfmNEnParallel{} \emph{parallel} items (one tool, several calls in one turn), \LfmNEnIrrel{} \emph{irrelevance} items (the offered tool does not fit, and the correct answer is no call) and \LfmNEnMulti{} \emph{multi-turn} items.
BFCL scores multi-turn conversations by executing calls in a stateful environment; we do not.
Instead we take the first two user turns, insert the gold first-turn calls with a placeholder tool response, and score only the second-turn calls, with every tool of the conversation offered as a distractor.
The Tajik version replaces the user turns with \LfmTranslateModel{} translations made with the same masking as the training data.
Tools, schemas and gold calls are identical across languages.
The translation step protects quoted text and links from translation, and \LfmTgUntranslated{} requests that are quoted as a whole were left in English (\LfmTgUntranslatedParallel{} of them parallel items). We found this while preparing this report and removed these items from both languages, so English and Tajik are scored on the same \LfmNAll{} items; every tool-calling number in this report, for every model, is computed on this cleaned set with the original scorer.
Table~\ref{tab:lfm-examples} in the appendix shows one item of each category.
No BFCL file is among the training sources, but we did not test for near-duplicates between Glaive and BFCL items.

\paragraph{Decoding and scoring.}
Every model is run with its chat template, the tools passed to the template, and greedy decoding.
The scorer parses the native LFM2.5 call syntax (a Python-style call list between two special tags, shown in Section~\ref{sec:release}) with Python's abstract-syntax-tree (\texttt{ast}) module, never by executing it, and matches predicted to gold calls by function name and argument overlap.
The main metric is \emph{full-call exact} accuracy over the \LfmNCall{} call-required items: an item counts only if every gold call is present, every argument key and value matches after light normalisation, and there are no extra calls or arguments.
We also report function-name accuracy, argument-key and argument-value accuracy, \emph{irrelevance} accuracy (no call emitted on the \LfmNIrrel{} no-call items), and the share of outputs that parse.

\paragraph{Lenient scoring.}
The strict parser only sees calls in the native format.
To separate a lost format from a lost skill, we re-scored all saved outputs with a lenient parser that also accepts JSON objects with a name and arguments, OpenAI-style function objects, untagged Python calls and XML-style calls, anywhere in the text, with the same matching rules.
A call must be a structured object; a function name mentioned in a sentence never counts.
The lenient parser reproduces the strict scores exactly on outputs in the native format, and a hand check of \LfmLenHandN{} random items where the two disagree found \LfmLenHandTrue{} true calls and no false positives.
A third variant, marked \emph{Inf.}, also credits a bare argument object to the single tool offered; we treat it as an upper bound.

\paragraph{General ability.}
To see what the two stages did to the model outside tool calling, we score every stage on three multiple-choice sets with four options each, using their full test sets: Belebele Tajik reading comprehension \citep{bandarkar2024belebele} (\LfmNBel{} items), the validation split of the Tajik Curated General Benchmark released with the Soro models \citep{liashkov2026soro} (\LfmNCur{} items, history, biology, literature, language, law and geography), and English MMLU \citep{hendrycks2021mmlu} (\LfmNMmlu{} items) as a check on English knowledge.
Each answer is chosen by the log-likelihood of the answer letter, without a chat template, for all stages alike.
These runs, the tool-calling run of the final CPT checkpoint and the re-run of the released model (Section~\ref{sec:release}) used one NVIDIA A10; all other tool-calling runs used H100s.

\paragraph{Statistics.}
Each cell is one training run and one decode.
We give 95\% percentile bootstrap intervals over items (10,000 resamples) and, for comparisons between two models on the same items, a paired bootstrap interval and an exact McNemar test (a binomial test on the items that only one of the two models gets right).
For a score of zero the bootstrap interval collapses to a point; the exact 97.5\% upper bound for zero of \LfmNCall{} items is \LfmZeroUpper\%.
We report $p$-values per comparison as a description and make no family-wise claim.
The intervals cover item sampling only, not variation between training seeds or between translations.
Decoding is not perfectly stable across GPU types either: re-running the released model on a different GPU changed \LfmRerunEnDiff{} of \LfmNAll{} English outputs and moved the English score by \LfmRerunEnShift{} points (Section~\ref{sec:release}). Differences of this size are within noise.

\section{Results}
\label{sec:results}

\begin{table*}[t]
\centering
\small
\begin{tabular}{lcc}
\toprule
Stage & English & Tajik \\
\midrule
Base & \LfmBaseEnExact{} [\LfmBaseEnExactLo{}, \LfmBaseEnExactHi{}] & \LfmBaseTgExact{} [\LfmBaseTgExactLo{}, \LfmBaseTgExactHi{}] \\
CPT epoch 1 & \LfmCptOneEnExact{} [\LfmCptOneEnExactLo{}, \LfmCptOneEnExactHi{}] & \LfmCptOneTgExact{} [\LfmCptOneTgExactLo{}, \LfmCptOneTgExactHi{}] \\
CPT epoch 2 & \LfmCptTwoEnExact{} [\LfmCptTwoEnExactLo{}, \LfmCptTwoEnExactHi{}] & \LfmCptTwoTgExact{} [\LfmCptTwoTgExactLo{}, \LfmCptTwoTgExactHi{}] \\
SFT epoch 1 (released) & \LfmSftOneEnExact{} [\LfmSftOneEnExactLo{}, \LfmSftOneEnExactHi{}] & \LfmSftOneTgExact{} [\LfmSftOneTgExactLo{}, \LfmSftOneTgExactHi{}] \\
SFT epoch 2 & \LfmSftTwoEnExact{} [\LfmSftTwoEnExactLo{}, \LfmSftTwoEnExactHi{}] & \LfmSftTwoTgExact{} [\LfmSftTwoTgExactLo{}, \LfmSftTwoTgExactHi{}] \\
\bottomrule
\end{tabular}
\caption{Strict full-call exact accuracy (\%) on the \LfmNCall{} call-required items of the \LfmNAll{}-item tool-calling test set, per language, with 95\% percentile bootstrap intervals (10,000 resamples of items). Greedy decoding. A zero score has a degenerate bootstrap interval; its exact 97.5\% upper bound is \LfmZeroUpper{}. Stages are the stock base, the CPT adapter after epoch 1 and 2 (merged), SFT epoch 1 on top of the final CPT model, and one more SFT epoch at half the learning rate on top of the merged epoch-1 model.}
\label{tab:lfm-main}
\end{table*}

\subsection{Main result}
Table~\ref{tab:lfm-main} gives strict full-call accuracy at every stage.
The base model is strong in English (\LfmBaseEnExact) and weak in Tajik (\LfmBaseTgExact).
After both CPT checkpoints the strict score is zero in both languages.
After one SFT epoch the model reaches \LfmSftOneEnExact{} in English and \LfmSftOneTgExact{} in Tajik.

\begin{figure}[t]
\centering
\includegraphics[width=\columnwidth]{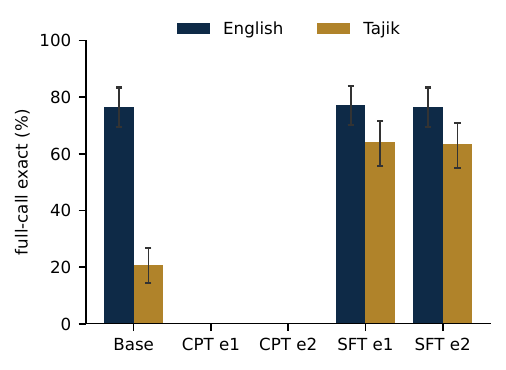}
\caption{Strict full-call exact accuracy by stage, English and Tajik, with 95\% bootstrap intervals.}
\label{fig:stages}
\end{figure}

Against the base, the Tajik gain is \LfmPairTgBaseSftDelta{} points (paired 95\% interval \LfmPairTgBaseSftLo{} to \LfmPairTgBaseSftHi; \LfmPairTgBaseSftOnlyB{} items fixed and \LfmPairTgBaseSftOnlyA{} broken; McNemar $p$ \LfmPairTgBaseSftP).
In English we detect no change (\LfmPairEnBaseSftDelta{} points, interval \LfmPairEnBaseSftLo{} to \LfmPairEnBaseSftHi); with \LfmNCall{} items, a change of a few points in either direction cannot be ruled out.
Figure~\ref{fig:stages} shows all stages together.
A gap of \LfmGapEnTgSft{} points between English and Tajik remains.

\begin{table*}[t]
\centering
\small
\begin{tabular}{lcccc}
\toprule
 & \multicolumn{2}{c}{English} & \multicolumn{2}{c}{Tajik} \\
\cmidrule(lr){2-3}\cmidrule(lr){4-5}
Category & Base & SFT e1 & Base & SFT e1 \\
\midrule
Single ($n=\LfmNEnSingle$) & \LfmBaseEnSingleExact{} [\LfmBaseEnSingleExactLo{}, \LfmBaseEnSingleExactHi{}] & \LfmSftOneEnSingleExact{} [\LfmSftOneEnSingleExactLo{}, \LfmSftOneEnSingleExactHi{}] & \LfmBaseTgSingleExact{} [\LfmBaseTgSingleExactLo{}, \LfmBaseTgSingleExactHi{}] & \LfmSftOneTgSingleExact{} [\LfmSftOneTgSingleExactLo{}, \LfmSftOneTgSingleExactHi{}] \\
Parallel ($n=\LfmNEnParallel$) & \LfmBaseEnParallelExact{} [\LfmBaseEnParallelExactLo{}, \LfmBaseEnParallelExactHi{}] & \LfmSftOneEnParallelExact{} [\LfmSftOneEnParallelExactLo{}, \LfmSftOneEnParallelExactHi{}] & \LfmBaseTgParallelExact{} [\LfmBaseTgParallelExactLo{}, \LfmBaseTgParallelExactHi{}] & \LfmSftOneTgParallelExact{} [\LfmSftOneTgParallelExactLo{}, \LfmSftOneTgParallelExactHi{}] \\
Multi-turn ($n=\LfmNEnMulti$) & \LfmBaseEnMultiExact{} [\LfmBaseEnMultiExactLo{}, \LfmBaseEnMultiExactHi{}] & \LfmSftOneEnMultiExact{} [\LfmSftOneEnMultiExactLo{}, \LfmSftOneEnMultiExactHi{}] & \LfmBaseTgMultiExact{} [\LfmBaseTgMultiExactLo{}, \LfmBaseTgMultiExactHi{}] & \LfmSftOneTgMultiExact{} [\LfmSftOneTgMultiExactLo{}, \LfmSftOneTgMultiExactHi{}] \\
\midrule
Overall full-call exact & \LfmBaseEnExact{} & \LfmSftOneEnExact{} & \LfmBaseTgExact{} & \LfmSftOneTgExact{} \\
Irrelevance accuracy ($n=\LfmNIrrel$) & \LfmBaseEnIrrel{} & \LfmSftOneEnIrrel{} & \LfmBaseTgIrrel{} & \LfmSftOneTgIrrel{} \\
\bottomrule
\end{tabular}
\caption{Per-category full-call exact accuracy (\%) for the stock base and the released model (SFT epoch 1), with 95\% bootstrap intervals; overall and irrelevance rows below. Intervals for small categories are wide. Greedy decoding.}
\label{tab:lfm-scope}
\end{table*}

\subsection{By category}
Table~\ref{tab:lfm-scope} splits the call-required items by category.
In Tajik the released model reaches \LfmSftOneTgSingleExact{} on single calls (base \LfmBaseTgSingleExact) and \LfmSftOneTgParallelExact{} on parallel calls (base \LfmBaseTgParallelExact).
Figure~\ref{fig:scope} in the appendix shows the same split as a chart.
Multi-turn items remain hard in both languages: \LfmSftOneEnMultiExact{} in English and \LfmSftOneTgMultiExact{} in Tajik, on only \LfmNEnMulti{} items each, so the intervals are wide.
Function-name accuracy after SFT is \LfmSftOneEnFn{} in English and \LfmSftOneTgFn{} in Tajik (Table~\ref{tab:lfm-support} in the appendix); most remaining Tajik errors are wrong argument values, not wrong functions.
Part of this gap is a property of the test set rather than of the model: the gold values stay in English, while the translated request may carry a value in Tajik (a city name in Cyrillic, for example), so a model that copies the value from the Tajik request is scored as wrong.
This is the argument-language mismatch described for translated BFCL by \citet{luo2026lostinexecution}; we have not measured how many of our errors it explains.

\subsection{What CPT removed: format and skill}
\label{sec:format}

\begin{table*}[t]
\centering
\small
\begin{tabular}{lrrrrrr}
\toprule
 & \multicolumn{3}{c}{English} & \multicolumn{3}{c}{Tajik} \\
\cmidrule(lr){2-4}\cmidrule(lr){5-7}
Stage & Tag & JSON & Other & Tag & JSON & Other \\
\midrule
Base & \LfmBaseEnTagCount{} & \LfmBaseEnJsonCount{} & \LfmBaseEnNeitherCount{} & \LfmBaseTgTagCount{} & \LfmBaseTgJsonCount{} & \LfmBaseTgNeitherCount{} \\
CPT epoch 1 & \LfmCptOneEnTagCount{} & \LfmCptOneEnJsonCount{} & \LfmCptOneEnNeitherCount{} & \LfmCptOneTgTagCount{} & \LfmCptOneTgJsonCount{} & \LfmCptOneTgNeitherCount{} \\
CPT epoch 2 & \LfmCptTwoEnTagCount{} & \LfmCptTwoEnJsonCount{} & \LfmCptTwoEnNeitherCount{} & \LfmCptTwoTgTagCount{} & \LfmCptTwoTgJsonCount{} & \LfmCptTwoTgNeitherCount{} \\
SFT epoch 1 (released) & \LfmSftOneEnTagCount{} & \LfmSftOneEnJsonCount{} & \LfmSftOneEnNeitherCount{} & \LfmSftOneTgTagCount{} & \LfmSftOneTgJsonCount{} & \LfmSftOneTgNeitherCount{} \\
\bottomrule
\end{tabular}
\caption{Output format of the generated text on all \LfmNAll{} items per language. Tag: contains the native tool-call start tag; JSON: no tag but a JSON-like name or function key; Other: neither. Greedy decoding.}
\label{tab:lfm-format}
\end{table*}

The zero after CPT is first a format failure.
Of \LfmNAll{} English outputs from the epoch-1 CPT checkpoint, \LfmCptOneEnTagCount{} contains the native call tag and \LfmCptOneEnJsonCount{} contain a JSON-like call instead (Table~\ref{tab:lfm-format}); in Tajik no output has the tag.
CPT on plain text taught the model to write Tajik but not to keep the special format that post-training had attached to tool calls.
The final CPT checkpoint (epoch \LfmCptEpochs), from which SFT starts, is no different: it emits \LfmToolCptFinEnTagCount{} tagged calls in English and \LfmToolCptFinTgTagCount{} in Tajik, and its strict score is \LfmToolCptFinEnExact{} in both languages. This checkpoint was scored later, on an A10 GPU, and with the strict parser only.

The lenient re-scoring (Table~\ref{tab:lfm-lenient}) shows the loss is not only a format problem.
Accepting any structured call, the epoch-1 checkpoint scores \LfmLenCptOneEnExact{} in English (base \LfmLenBaseEnExact) and \LfmLenCptOneTgExact{} in Tajik (base \LfmLenBaseTgExact), and the epoch-2 checkpoint scores lower again (\LfmLenCptTwoEnExact{} and \LfmLenCptTwoTgExact).
Multi-turn accuracy, the only category where the right function cannot be copied from a single offered tool, falls to \LfmLenCptOneEnMultiExact{} in English and \LfmLenCptOneTgMultiExact{} in Tajik after one epoch.
Even the most generous variant (Inf.) leaves the CPT checkpoints far below the base.
The skill that the strict scorer hid was real but small; in English it shrank further in the second CPT epoch, while in Tajik it stayed at the same low level.

The lenient scorer also corrects the base model.
In Tajik the base often answers with its own XML-style or JSON calls, which the strict scorer counts as wrong or as no call.
Its lenient Tajik accuracy is \LfmLenBaseTgExact{} rather than \LfmBaseTgExact, and its Tajik irrelevance accuracy is \LfmLenBaseTgIrrel{} rather than \LfmBaseTgIrrel.
Against this fairer baseline the released model's Tajik gain is still \LfmLenSftOneTgExact{} against \LfmLenBaseTgExact, and the released model's outputs score the same under both parsers.

\begin{table*}[t]
\centering
\small
\footnotesize
\setlength{\tabcolsep}{3pt}
\begin{tabular}{lrrrrrrrrrr}
\toprule
 & \multicolumn{5}{c}{English} & \multicolumn{5}{c}{Tajik} \\
\cmidrule(lr){2-6}\cmidrule(lr){7-11}
Stage & \multicolumn{3}{c}{Full-call exact} & \multicolumn{2}{c}{Function name} & \multicolumn{3}{c}{Full-call exact} & \multicolumn{2}{c}{Function name} \\
\cmidrule(lr){2-4}\cmidrule(lr){5-6}\cmidrule(lr){7-9}\cmidrule(lr){10-11}
 & Str. & Len. & Inf. & Str. & Len. & Str. & Len. & Inf. & Str. & Len. \\
\midrule
Base & \LfmBaseEnExact{} & \LfmLenBaseEnExact{} & \LfmInfBaseEnExact{} & \LfmBaseEnFn{} & \LfmLenBaseEnFn{} & \LfmBaseTgExact{} & \LfmLenBaseTgExact{} & \LfmInfBaseTgExact{} & \LfmBaseTgFn{} & \LfmLenBaseTgFn{} \\
CPT epoch 1 & \LfmCptOneEnExact{} & \LfmLenCptOneEnExact{} & \LfmInfCptOneEnExact{} & \LfmCptOneEnFn{} & \LfmLenCptOneEnFn{} & \LfmCptOneTgExact{} & \LfmLenCptOneTgExact{} & \LfmInfCptOneTgExact{} & \LfmCptOneTgFn{} & \LfmLenCptOneTgFn{} \\
CPT epoch 2 & \LfmCptTwoEnExact{} & \LfmLenCptTwoEnExact{} & \LfmInfCptTwoEnExact{} & \LfmCptTwoEnFn{} & \LfmLenCptTwoEnFn{} & \LfmCptTwoTgExact{} & \LfmLenCptTwoTgExact{} & \LfmInfCptTwoTgExact{} & \LfmCptTwoTgFn{} & \LfmLenCptTwoTgFn{} \\
SFT epoch 1 (released) & \LfmSftOneEnExact{} & \LfmLenSftOneEnExact{} & \LfmInfSftOneEnExact{} & \LfmSftOneEnFn{} & \LfmLenSftOneEnFn{} & \LfmSftOneTgExact{} & \LfmLenSftOneTgExact{} & \LfmInfSftOneTgExact{} & \LfmSftOneTgFn{} & \LfmLenSftOneTgFn{} \\
\bottomrule
\end{tabular}
\caption{Strict versus lenient scoring of the same saved predictions (\LfmNCall{} call-required items per language, percent). Str.: strict scorer, native tag format only. Len.: any structured call (tagged, untagged Python, or JSON with a function name and arguments). Inf.: Len. plus JSON objects that carry arguments but no function name, attributed to the offered tool on single-tool items. Matching and value normalisation are identical in all columns. SFT epoch 2 has no saved generated text, so it cannot be re-scored; its strict per-item correctness comes from the scorer's error file.}
\label{tab:lfm-lenient}
\end{table*}

\begin{table*}[t]
\centering
\small
\begin{tabular}{lccc}
\toprule
Stage & Belebele (tg) & Tajik curated & MMLU (en) \\
\midrule
Base & \LfmMcqBaseBel{} [\LfmMcqBaseBelLo{}, \LfmMcqBaseBelHi{}] & \LfmMcqBaseCur{} [\LfmMcqBaseCurLo{}, \LfmMcqBaseCurHi{}] & \LfmMcqBaseMmlu{} [\LfmMcqBaseMmluLo{}, \LfmMcqBaseMmluHi{}] \\
CPT final (epoch 3) & \LfmMcqCptFinBel{} [\LfmMcqCptFinBelLo{}, \LfmMcqCptFinBelHi{}] & \LfmMcqCptFinCur{} [\LfmMcqCptFinCurLo{}, \LfmMcqCptFinCurHi{}] & \LfmMcqCptFinMmlu{} [\LfmMcqCptFinMmluLo{}, \LfmMcqCptFinMmluHi{}] \\
SFT epoch 1 (released) & \LfmMcqSftOneBel{} [\LfmMcqSftOneBelLo{}, \LfmMcqSftOneBelHi{}] & \LfmMcqSftOneCur{} [\LfmMcqSftOneCurLo{}, \LfmMcqSftOneCurHi{}] & \LfmMcqSftOneMmlu{} [\LfmMcqSftOneMmluLo{}, \LfmMcqSftOneMmluHi{}] \\
SFT epoch 2 & \LfmMcqSftTwoBel{} [\LfmMcqSftTwoBelLo{}, \LfmMcqSftTwoBelHi{}] & \LfmMcqSftTwoCur{} [\LfmMcqSftTwoCurLo{}, \LfmMcqSftTwoCurHi{}] & \LfmMcqSftTwoMmlu{} [\LfmMcqSftTwoMmluLo{}, \LfmMcqSftTwoMmluHi{}] \\
\bottomrule
\end{tabular}
\caption{General ability by stage: accuracy (\%) with 95\% percentile bootstrap intervals. Belebele Tajik reading comprehension (\LfmNBel{} items), a Tajik multiple-choice knowledge set (\LfmNCur{} items) and English MMLU (\LfmNMmlu{} items), full test sets, four options each. Scored by the log-likelihood of the answer letter, without a chat template, for every stage. All rows in this table were run on one A10 GPU.}
\label{tab:lfm-general}
\end{table*}

\subsection{General Tajik and English ability}
\label{sec:general}
Table~\ref{tab:lfm-general} shows the cost and the gain outside tool calling.
On the Tajik knowledge set the released model scores \LfmMcqSftOneCur{} against \LfmMcqBaseCur{} for the base (paired difference \LfmMcqPairCurBaseSftDelta{} points, interval \LfmMcqPairCurBaseSftLo{} to \LfmMcqPairCurBaseSftHi; $p$ \LfmMcqPairCurBaseSftP), and most of this gain appears already after CPT (\LfmMcqCptFinCur).
On Belebele Tajik the change from base to released model is small and not clearly different from zero (\LfmMcqPairBelBaseSftDelta{} points, interval \LfmMcqPairBelBaseSftLo{} to \LfmMcqPairBelBaseSftHi).
English knowledge pays for it: MMLU falls from \LfmMcqBaseMmlu{} to \LfmMcqCptFinMmlu{} after CPT (\LfmMcqPairMmluBaseCptDelta{} points) and recovers only to \LfmMcqSftOneMmlu{} after SFT.
In absolute terms the model's general Tajik remains weak: on four-option questions, where guessing scores about a quarter, the released model is only \LfmMcqSftOneCurAboveChance{} points above chance on the knowledge benchmark and \LfmMcqSftOneBelAboveChance{} on Belebele.
\ModelName{} is a tool-calling model; it should not be used as a general Tajik assistant.

\subsection{Weaknesses}
\label{sec:weak}

\paragraph{Abstention.}
After SFT the model calls a tool even when none fits.
Irrelevance accuracy falls from \LfmBaseEnIrrel{} to \LfmSftOneEnIrrel{} in English (paired $p$ \LfmPairEnIrrelBaseSftP) and to \LfmSftOneTgIrrel{} in Tajik.
Measured with the lenient parser, the Tajik base abstained on \LfmLenBaseTgIrrel\%, not \LfmBaseTgIrrel\%, so the real Tajik drop is smaller than the strict numbers suggest, but it is still a drop.
We believe the cause is in our data preparation: the filter that converted the tool-calling sources kept only conversations with at least one call, so no SFT conversation offered tools and then answered without one. In Glaive alone this dropped \LfmGlaiveNoCall{} such conversations (of \LfmGlaiveToolsOffered{} that offer tools). We have not yet retrained with them, so the cause is likely but not tested.

\paragraph{Multi-turn.}
Multi-turn accuracy is low before and after SFT.
Our multi-turn items are a simplified form of BFCL's, without a real environment, so these numbers say little about long agent sessions.

\paragraph{Continued SFT.}
Training one more epoch from the merged epoch-1 model at half the learning rate changes English by \LfmPairEnSftOneTwoDelta{} and Tajik by \LfmPairTgSftOneTwoDelta{} points (\LfmPairTgSftOneTwoOnlyA{} items broken and \LfmPairTgSftOneTwoOnlyB{} fixed in Tajik; $p$ \LfmPairTgSftOneTwoP), and multi-turn accuracy does not move in either language.
These differences are smaller than the GPU-type noise we measured, so we see no evidence that more training on this mix fixes the weaknesses above; we release the epoch-1 model, which was already the release candidate before this run.

\section{Discussion}
\label{sec:discussion}

\paragraph{Why we used CPT before SFT.}
Our other work found that on bases that already handle Tajik, CPT before SFT can do more harm than good.
LFM2.5-2.6B is not such a base: before training it handles a Tajik request poorly even when the tools and gold calls are in English.
For a base like this we expected a stage of plain Tajik text to be needed, and we accepted that it costs the tool format and part of the skill.
We did not train an SFT-only arm, so we cannot show that CPT was necessary; the Tajik knowledge gain after CPT (Table~\ref{tab:lfm-general}) is the only direct evidence that it taught the model something.

\paragraph{SFT brought the skill back.}
The SFT mix rebuilt the English score to the base level and lifted Tajik far above it in one epoch.
We did not train an SFT arm without English replay, so we cannot say how much of the recovery is due to replay rather than to the translated Tajik calls alone.

\paragraph{Practical advice.}
From this single run, one model and one training seed, we would offer five suggestions to anyone adapting a small tool-calling model to a new language:
\begin{enumerate}
\item Check the tokenizer for every letter of the target alphabet before training.
\item Score tool calls with both a strict and a lenient parser. The strict score is what a deployment sees; the lenient one tells you whether a zero is a lost format or a lost skill. Without it, the base model in our study would have looked worse at Tajik tool calling and better at abstaining than it is.
\item Expect raw-text CPT to remove tool calling, and plan the SFT stage to rebuild it with replay in the original language.
\item Put no-call examples into the SFT data on purpose, and check that data filters do not remove them; ours did, and the model now almost always calls.
\item Measure general knowledge in the original language as well. After SFT our model still scores \LfmMcqBaseMmlu{} against \LfmMcqSftOneMmlu{} on English MMLU for the base, a loss that tool-calling scores alone did not show.
\end{enumerate}

\section{Release}
\label{sec:release}

\paragraph{What we release.}
The merged weights of \ModelName{} (SFT epoch 1) are public on the Hugging Face Hub as \texttt{Tohirju/Saidzoda-Tajik-LFM-2.6B} under the licence below; access is granted on request through the Hub.
We verified that the published file is the one we evaluated by comparing SHA-256 hashes, and re-ran the tool-calling test on the file downloaded from the Hub, on a different GPU type (A10 instead of H100).
The Tajik score is identical (\LfmToolRerunTgExact) and \LfmRerunTgSame{} of \LfmNAll{} Tajik outputs are byte-identical; in English \LfmRerunEnSame{} outputs are identical and the score moves from \LfmSftOneEnExact{} to \LfmToolRerunEnExact, a net difference of two items that we attribute to numerical differences between GPU types.
The evaluation set, the scorer, the training configurations and the evidence pack that produces every number in this report are available on request; Table~\ref{tab:lfm-examples} shows example items.
This is the first model our lab releases with a technical report.
Our earlier Tajik and Central Asian models, including other tool-calling models, are on the Hub without reports; we plan to describe them in reports of the same kind, starting with the Qwen3 series.

\paragraph{How to run it.}
The repository holds merged weights in bfloat16 (about five gigabytes), the tokenizer and the chat template; no adapters are needed.
Tools are passed to the chat template in the usual JSON-schema form, and the model answers in the native LFM2.5 syntax, which the caller parses and executes, for example:
\begin{center}\footnotesize\texttt{<|tool\_call\_start|>}\\\texttt{[get\_weather(city='Dushanbe')]}\\\texttt{<|tool\_call\_end|>}\end{center}
We evaluated only greedy decoding.

\paragraph{Licence.}
The base model is distributed under the LFM Open License v1.0, and \ModelName{} is a derivative work under that licence.
We ship the licence text with the weights, mark the files we changed, and keep the original notices.
Commercial use by an organisation with annual revenue at or above the licence threshold is not covered by the base licence.

\paragraph{Intended use.}
The model is meant for turning Tajik requests into calls to tools that the developer defines, in settings where a person or a validation layer checks a call before it changes anything.
It should not be used where a wrong call causes harm without review, and it should not be relied on to decline requests it cannot serve.

\section{Conclusion}

We adapted a 2.6B hybrid model with almost no Tajik ability into a Tajik tool-calling model with two training stages and about \LfmCptRuntimeHours{} hours of continued pretraining on three GPUs.
Strict Tajik tool-calling accuracy rose from \LfmBaseTgExact{} to \LfmSftOneTgExact{} with no detectable change in English, and Tajik knowledge rose while English MMLU fell.
Continued pretraining on plain text removed the tool-call format and part of the skill, and one epoch of SFT with English replay restored it.
The model is weak at abstaining and at multi-turn calls; preference or reinforcement training aimed at these two weaknesses is our next step.

\section*{Limitations}

\begin{itemize}
\item \textbf{General ability is measured only by multiple choice.} We score reading comprehension and knowledge by answer log-likelihood; we have no perplexity, generation or human evaluation of the model's free Tajik text, and make no claim about its quality.
\item \textbf{Machine-translated test set.} The Tajik user turns were translated by the same system that translated the training data, and no native speaker has checked them. Tajik written by real users will differ in style and spelling, and the scores may be optimistic. \LfmTgUntranslated{} Tajik requests that the translation step left in English were removed from the test set in both languages.
\item \textbf{Small test set and one run.} \LfmNCall{} call-required items, \LfmNIrrel{} no-call items and \LfmNEnMulti{} multi-turn items per language, one training seed and greedy decoding. Our intervals do not cover seed or translation variation.
\item \textbf{Simplified multi-turn.} We do not execute calls in an environment.
\item \textbf{Missing arms.} We trained no SFT arm without replay, no arm without CPT, and no arm with the dropped no-call conversations restored. The final CPT checkpoint was scored later than the others and on a different GPU type, and was not re-scored with the lenient parser.
\item \textbf{Corpus defect.} Between \LfmGarbledLoPct{} and \LfmGarbledHiPct\% of the CPT documents are mis-decoded text (Section~\ref{sec:recipe}). We did not inspect them or measure their effect.
\item \textbf{Possible overlap.} We did not test for near-duplicates between the Glaive training data and the BFCL test items.
\item \textbf{No deployment measurements.} We report no latency, memory or quantised results.
\item \textbf{One run.} Our suggestions in Section~\ref{sec:discussion} come from one model, one seed and a corpus with a known defect; we did not tune the LoRA rank or the learning rate.
\item \textbf{Ethics and risks.} The web text was filtered for script and encoding, not for personal data, toxicity or bias, and we did not evaluate the model for any of these. Tajik web text leans towards news and official sources, so other varieties and registers are less covered. A model that calls tools when it should not can take wrong actions; it should only run behind a human or an automatic check of every call.
\end{itemize}

\section*{Use of AI Assistants}

We used AI assistants (Anthropic Claude models) to write data-processing and analysis scripts, to search for and check references, and to draft parts of the text.
The authors designed the study, ran the training, checked every number against the recorded files, and take full responsibility for the content.
Every number in this report is generated by a script from logged files, and each reference was checked against its arXiv or DOI record.
AI assistants are not authors.

\section*{Acknowledgements}

We thank NVIDIA for access to the NVIDIA Inception programme, through which the training compute for this work was provided.

\bibliography{refs}
\bibliographystyle{acl_natbib}

\appendix
\section{Training and Data Details}
\label{app:details}

\begin{table*}[t]
\centering
\small
\begin{tabular}{lccc}
\toprule
Setting & CPT & SFT epoch 1 & SFT epoch 2 \\
\midrule
Starting weights & LFM2.5-2.6B-Base & CPT merged & SFT e1 merged \\
Adapter & LoRA, rsLoRA & LoRA, rsLoRA & LoRA, rsLoRA \\
LoRA rank / alpha & \LfmCptLoraR{} / \LfmCptLoraAlpha{} & \LfmSftLoraR{} / \LfmSftLoraAlpha{} & \LfmSftLoraR{} / \LfmSftLoraAlpha{} \\
Target modules & \multicolumn{3}{c}{q\_proj, k\_proj, v\_proj, out\_proj, in\_proj, w1, w2, w3} \\
Sequence length & \LfmCptSeqLen{} & \LfmSftSeqLen{} & \LfmSftSeqLen{} \\
Sample packing & \LfmCptPacking{} & \LfmSftPacking{} & \LfmSftPacking{} \\
Epochs & \LfmCptEpochs{} & \LfmSftOneEpochs{} & \LfmSftTwoEpochs{} \\
Optimizer steps & \LfmCptSteps{} & \LfmSftOneSteps{} & \LfmSftTwoSteps{} \\
Peak learning rate & \LfmCptLr{} & \LfmSftOneLr{} & \LfmSftTwoLr{} \\
Schedule / warmup ratio & \LfmCptSched{} / \LfmCptWarmup{} & \LfmSftSched{} / \LfmSftWarmup{} & \LfmSftSched{} / \LfmSftWarmup{} \\
Grad-norm clip & \LfmCptClip{} & \LfmSftClip{} & \LfmSftClip{} \\
Micro batch / grad accum / GPUs & \LfmCptMicroBatch{} / \LfmCptGradAccum{} / \LfmCptGpus{} & \LfmSftMicroBatch{} / \LfmSftGradAccum{} / \LfmSftGpus{} & \LfmSftMicroBatch{} / \LfmSftGradAccum{} / \LfmSftGpus{} \\
Effective batch (sequences) & \LfmCptEffBatch{} & \LfmSftEffBatch{} & \LfmSftEffBatch{} \\
Wall-clock (hours, H100) & \LfmCptRuntimeHours{} & \LfmSftOneRuntimeHours{} & \LfmSftTwoRuntimeHours{} \\
First / last step loss & \LfmCptLossFirst{} / \LfmCptLossLast{} & \LfmSftOneLossFirst{} / \LfmSftOneLossLast{} & \LfmSftTwoLossFirst{} / \LfmSftTwoLossLast{} \\
Mean train loss & \LfmCptTrainLoss{} & \LfmSftOneTrainLoss{} & \LfmSftTwoTrainLoss{} \\
\bottomrule
\end{tabular}
\caption{Training recipe, read from the axolotl configs and training logs. CPT is raw-text completion on the cleaned Tajik corpus; SFT is chat-template tool-calling SFT with English replay and Tajik instructions on top of the merged CPT model; SFT epoch 2 continues from the merged SFT epoch 1 model at a lower rate. Losses are single-step values and mean train loss is the trainer summary.}
\label{tab:lfm-recipe}
\end{table*}

\begin{table*}[t]
\centering
\small
\begin{tabular}{lrr}
\toprule
CPT source & Documents & Share (\%) \\
\midrule
HPLT 2.0 cleaned & \LfmSrcDocsHplt{} & \LfmSrcPctHplt{} \\
FineWeb-2 & \LfmSrcDocsFineweb{} & \LfmSrcPctFineweb{} \\
MADLAD-400 & \LfmSrcDocsMadlad{} & \LfmSrcPctMadlad{} \\
C4 & \LfmSrcDocsCfour{} & \LfmSrcPctCfour{} \\
Tajik Wikipedia corpus & \LfmSrcDocsWiki{} & \LfmSrcPctWiki{} \\
Shahnameh corpus & \LfmSrcDocsShah{} & \LfmSrcPctShah{} \\
Tajik\_Pretrain\_370M (packed rows) & \LfmSrcDocsPretrain{} & \LfmSrcPctPretrain{} \\
\midrule
Pre-clean corpus & \LfmDocsPre{} & 100.0 \\
Dropped, residual legacy encoding & \LfmDroppedResidual{} &  \\
Dropped, exact duplicates after cleaning & \LfmDroppedDup{} &  \\
Kept for CPT & \LfmDocsKept{} &  \\
\bottomrule
\end{tabular}
\caption{CPT corpus before cleaning, by source, and the cleaning outcome. Shares are of the pre-clean document count (\LfmDocsPre{} documents, \LfmTokensPre{} LFM2 tokens). Cleaning removed documents with residual legacy-font characters and ran on \LfmDocsInput{} input documents (the cleaning log counts two exact duplicates that the combine step had already removed), so the pre-clean row minus the dropped rows does not equal the kept row; document counts per source after cleaning were not recorded.}
\label{tab:lfm-cptdata}
\end{table*}

\begin{table*}[t]
\centering
\small
\begin{tabular}{lrr}
\toprule
SFT component & Rows & Share (\%) \\
\midrule
glaive-function-calling-v2 & \LfmToolRowsGlaive{} &  \\
hermes-function-calling-v1, func\_calling & \LfmToolRowsHermesFc{} &  \\
hermes-function-calling-v1, func\_calling\_singleturn & \LfmToolRowsHermesSingle{} &  \\
hermes-function-calling-v1, glaive\_func\_calling & \LfmToolRowsHermesGlaive{} &  \\
ToolACE & \LfmToolRowsToolAce{} &  \\
Tool-call rows after dedup & \LfmToolRowsTotal{} &  \\
\midrule
Tajik tool calling (all \LfmToolRowsTotal{} rows translated) & \LfmMixTgTool{} & \LfmMixTgPct{} \\
English tool calling (replay) & \LfmMixEnTool{} & \LfmMixEnPct{} \\
Tajik instruction chat & \LfmMixInstruct{} & \LfmMixInsPct{} \\
SFT mix & \LfmSftMixRows{} & 100 \\
Trained rows (at most \LfmSftSeqLen{} tokens) & \LfmSftPretokRows{} &  \\
\bottomrule
\end{tabular}
\caption{Tool-call source rows (top) and the SFT mix (bottom). Mix counts follow the ratio arithmetic of the mix script, whose printed counts were not logged; they sum to the logged mix size. \LfmSurfaceStrings{} surface strings were machine-translated to Tajik for the tool-calling component.}
\label{tab:lfm-sftdata}
\end{table*}

\begin{table*}[t]
\centering
\small
\begin{tabular}{llrrrrrrr}
\toprule
Stage & Lang & Fn name & Arg key & Arg value & Full call & Irrel. & Valid parse & Extra call \\
\midrule
Base & EN & \LfmBaseEnFn{} & \LfmBaseEnArgKey{} & \LfmBaseEnArgVal{} & \LfmBaseEnExact{} & \LfmBaseEnIrrel{} & \LfmBaseEnParse{} & \LfmBaseEnExtra{} \\
 & TG & \LfmBaseTgFn{} & \LfmBaseTgArgKey{} & \LfmBaseTgArgVal{} & \LfmBaseTgExact{} & \LfmBaseTgIrrel{} & \LfmBaseTgParse{} & \LfmBaseTgExtra{} \\
\midrule
SFT epoch 1 (released) & EN & \LfmSftOneEnFn{} & \LfmSftOneEnArgKey{} & \LfmSftOneEnArgVal{} & \LfmSftOneEnExact{} & \LfmSftOneEnIrrel{} & \LfmSftOneEnParse{} & \LfmSftOneEnExtra{} \\
 & TG & \LfmSftOneTgFn{} & \LfmSftOneTgArgKey{} & \LfmSftOneTgArgVal{} & \LfmSftOneTgExact{} & \LfmSftOneTgIrrel{} & \LfmSftOneTgParse{} & \LfmSftOneTgExtra{} \\
\midrule
SFT epoch 2 & EN & \LfmSftTwoEnFn{} & \LfmSftTwoEnArgKey{} & \LfmSftTwoEnArgVal{} & \LfmSftTwoEnExact{} & \LfmSftTwoEnIrrel{} & \LfmSftTwoEnParse{} & \LfmSftTwoEnExtra{} \\
 & TG & \LfmSftTwoTgFn{} & \LfmSftTwoTgArgKey{} & \LfmSftTwoTgArgVal{} & \LfmSftTwoTgExact{} & \LfmSftTwoTgIrrel{} & \LfmSftTwoTgParse{} & \LfmSftTwoTgExtra{} \\
\bottomrule
\end{tabular}
\caption{Supporting metrics (\%) for the stock base, the released model (SFT e1) and SFT epoch 2. Fn name, arg key and arg value are micro-averaged over gold calls and keys; full call is the strict item-level score; Irrel. is the share of the \LfmNIrrel{} no-call items answered without a tool call; Valid parse is the share of emitted call blocks that parse; Extra call is the share of call-required items with a surplus predicted call. Greedy decoding.}
\label{tab:lfm-support}
\end{table*}

\begin{table*}[t]
\centering
\small
\footnotesize
\begin{tabular}{p{1.7cm}p{5.4cm}p{3.6cm}p{4.6cm}}
\toprule
Category & Tajik request & Offered tools & Gold call \\
\midrule
Single & \tj{Ба ман дар бораи филми} 'Interstellar' \tj{маълумот диҳед} & \texttt{movie\_details.brief} & \texttt{movie\_details.brief(\allowbreak{}title='Interstellar')} \\
Parallel & \tj{Агар ман сафари худро аз Ню-Йорк сар карда, ба Лос-Анҷелес, сипас аз Лос-Анҷелес ба Майами ва дар ниҳоят аз Майами ба Ню-Йорк равам, масофаи умумии километрҳо чӣ гуна аст}? & \texttt{geo\_distance.calculate} & \texttt{geo\_distance.calculate(\allowbreak{}start\_location='New York',\allowbreak{} end\_location='Los Angeles',\allowbreak{} units='kilometers');\allowbreak{} geo\_distance.calculate(\allowbreak{}start\_location='Los Angeles',\allowbreak{} end\_location='Miami',\allowbreak{} units='kilometers');\allowbreak{} geo\_distance.calculate(\allowbreak{}start\_location='Miami',\allowbreak{} end\_location='New York',\allowbreak{} units='kilometers')} \\
Irrelevance & \tj{Оё шумо метавонед филми хуберо барои тамошо тавсия диҳед}? & \texttt{word\_count} & \texttt{(\allowbreak{}no call)} \\
Multi-turn (2nd turn) & \tj{Азбаски мо ҳама тайёрем, оё шумо метавонед зуд фишори чархҳоро тафтиш кунед, то боварӣ ҳосил кунед, ки ҳама чиз барои сафари мо бехатар аст}? & 31 tools, including \texttt{check\_tire\_pressure} & \texttt{check\_tire\_pressure(\allowbreak{})} \\
\bottomrule
\end{tabular}
\caption{Example items from the Tajik tool-calling test set, the first item of each category whose request was translated. Tools and gold calls are the original BFCL ones in English. The full set and the scorer are available on request.}
\label{tab:lfm-examples}
\end{table*}

\begin{figure}[t]
\centering
\includegraphics[width=\columnwidth]{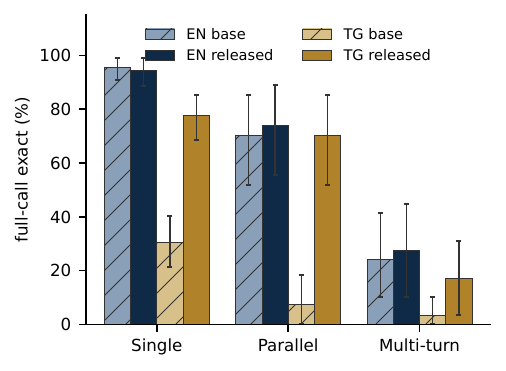}
\caption{Strict full-call exact accuracy by category for the base and released models.}
\label{fig:scope}
\end{figure}

\end{document}